\PassOptionsToPackage{backref=page}{hyperref}
\documentclass[11pt]{article}

\newif\ifauthordecided
\authordecidedtrue

\newif\ifarxiv
\arxivtrue

\newif\ifcompact
\compacttrue %

\newif\ifperfect
\perfectfalse %

\ifarxiv
    \usepackage{acl}
\else 
    \usepackage[review]{acl}
\fi

\usepackage{times}
\usepackage{latexsym}

\usepackage{tabularx}

\usepackage[T1]{fontenc}

\usepackage[utf8]{inputenc}

\usepackage{microtype}
\usepackage{sec/package}

\usepackage{comment}

\title{
Do LLMs Think Fast and Slow? 
A Causal Study on Sentiment Analysis

}

\author{Zhiheng Lyu\thanks{Equal contributions.}
\\
  University of Hong Kong \\
  \texttt{cogito@connect.hku.hk
  } \\\And
  Zhijing Jin\samethanks \\
  MPI \& University of Toronto \\
  \texttt{zjin@cs.toronto.edu} \\\And
  Fernando Gonzalez \\
  ETH Zürich \\
  \texttt{fer.adauto@gmail.com} \\\AND
  Rada Mihalcea \\
  University of Michigan \\
  \texttt{mihalcea@umich.edu} \\\And
  Bernhard Sch\"olkopf \\
  MPI for Intelligent Systems \\
  \texttt{bs@tue.mpg.de}\\\And
  Mrinmaya Sachan \\
  ETH Zürich \\
  \texttt{msachan@ethz.ch} \\
}

\begin{document}\maketitle

\begin{abstract}
Sentiment analysis (SA) aims to identify the sentiment expressed in a text, such as a product review. Given a review and the sentiment associated with it, this work formulates SA as a combination of two tasks: (1) a causal discovery task that distinguishes whether a review ``primes'' the sentiment (Causal Hypothesis C1), or the sentiment ``primes'' the review (Causal Hypothesis C2); and (2) the traditional prediction task to model the sentiment using the review as input.
Using the peak-end rule in psychology, we classify a sample as C1 if its overall sentiment score approximates an average of all the sentence-level sentiments in the review, and C2 if the overall sentiment score approximates an average of the peak and end sentiments.
For the prediction task, we use the discovered causal mechanisms behind the samples to improve LLM performance by proposing \textit{causal prompts} that give the models an inductive bias of the underlying causal graph, leading to {substantial improvements} by up to 32.13 F1 points on zero-shot five-class SA.\footnote{Our code 
\ifarxiv
is at  \href{https://github.com/cogito233/causal-sa}{https://github.com/cogito233/causal-sa}.
\fi
}
\end{abstract}

\section{Introduction}
Sentiment analysis (SA) is the task of identifying the sentiment $y$ given a piece of text $x$.
The field has a rich history originating from subjectivity analysis \cite{wiebe1994tracking,hatzivassiloglou2000effects}, and developed rapidly with the availability of large opinionated online data such as reviews with ratings \cite[\textit{inter alia}]{turney2002thumbs,nasukawa2003sentiment,zhang2015character,keung-etal-2020-multilingual}. 

\begin{figure*}[t]
    \centering
    \includegraphics[width=\textwidth]{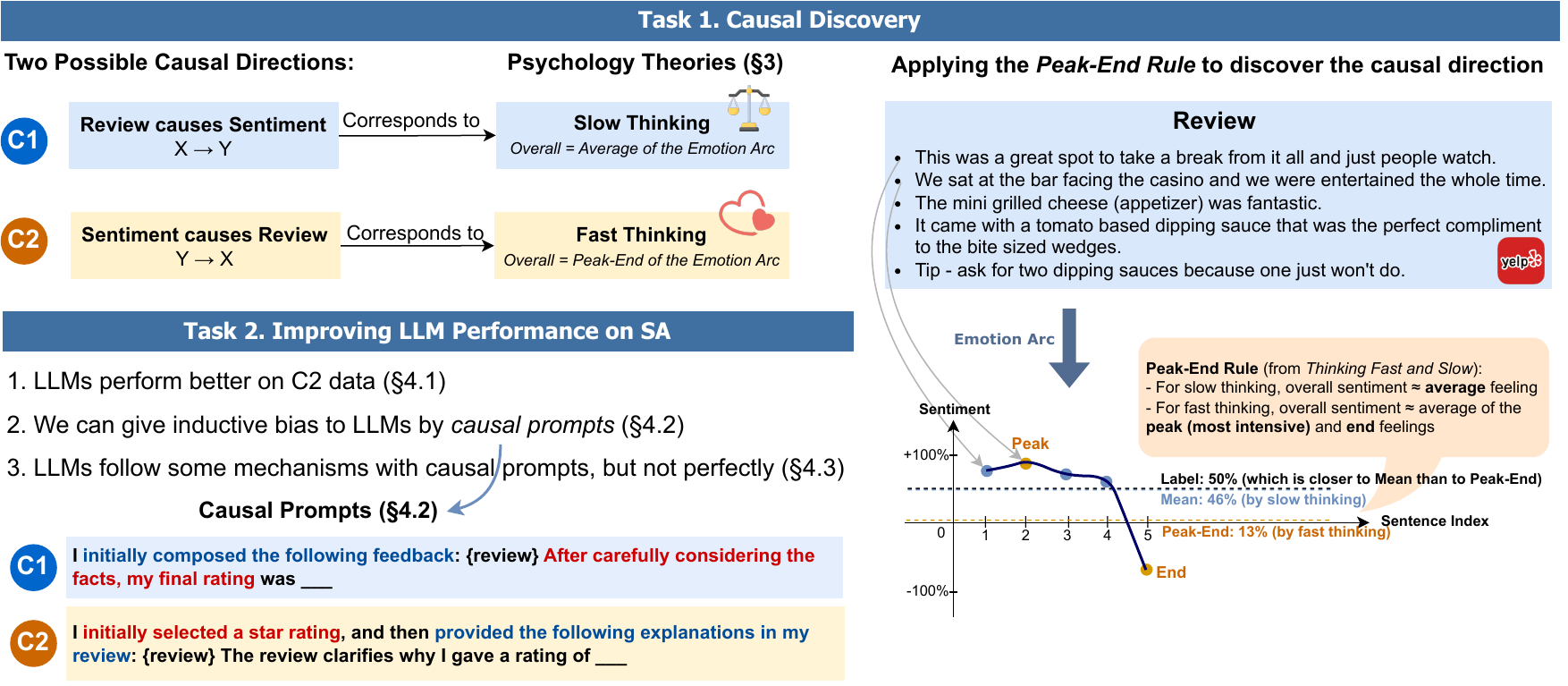}
    \caption{
    An overview of the paper structure, where we first investigate the causal discovery task, and then use it to improve LLM performance. For each document-level text review, we parse its \textit{emotion arc} consisting of the sentiment of each sentence in the review, and then use the peak-end rule \cite{kahneman1993more,kahneman2011thinking} to identify whether the overall sentiment is an average of the arc (corresponding to \textit{Slow Thinking}), or an average of the peak and end sentiments (corresponding to \textit{Fast Thinking}).
    }
    \label{fig:structure}
\end{figure*}

Despite recent advances in large language models (LLMs), 
it is still challenging to address the fine-grained five-class SA (which corresponds to the five star ratings in most datasets) for document-level classification \cite{Choi_2020,fei-etal-2023-reasoning,math9212722}, due to the subtle nature of the task including aspects such as inter-aspect relations, commonsense reasoning, 
among others 
\cite{poria2023beneath,venkit-etal-2023-sentiment}.

In this paper, we propose a causally-informed solution for the SA task. Different from the approach of naïvely applying up-to-date LLMs, we leverage insights from causal inference to
propose a reformulation for SA into two tasks, as in \cref{fig:structure}: (1) a causal discovery task to identify the cause-effect relation between the review $X$ and the sentiment $Y$, and (2) the traditional prediction task $f: {x} \mapsto y$ to model the sentiment using the review as input.

We first look into the causal discovery task.
In the study of affect science \cite{salovey2004emotional,barrett2006solving,feinstein2013lesion}, language can be the cause of emotion \cite{satpute2013functional,kassam2013effects} -- namely a review priming the following sentiment, i.e., the Causal Hypothesis C1 of $X\rightarrow Y$); or emotion can affect the use of language \cite{barrett2006solving} -- namely sentiment priming the review as an ad-hoc justification for the emotion, i.e., the Causal Hypothesis C2 of $Y\rightarrow X$.
These two processes might arise from the data annotation process \cite{jin-etal-2021-causal}, but is hard to discover post-hoc in existing datasets.

Given the possibility of both causal directions $X\rightarrow Y$ or $Y\rightarrow X$ in the SA data, we identify the actual underlying mechanism based on insights from psychology \cite{kahneman2011thinking,epstein1994integration}. Specifically, we identify the correspondence of the above two causal mechanisms with the \textit{Fast} and \textit{Slow Thinking} systems \cite{kahneman2011thinking}: (1) a review-driven sentiment (as in C1) largely resembles the Slow Thinking process applying reasoning based on evidence, and (2) the process of first coming up with the sentiment and then justifying it by a review (as in C2) conforms to Fast Thinking.
Given this correspondence, we apply the peak-end rule from psychology \cite{kahneman1993more,kahneman2011thinking}. As shown in the right part of \cref{fig:structure}, we classify a sample as C1 if its overall sentiment score approximates an average of all the sentence-level sentiments in the review, and as C2 if the overall sentiment score approximates an average of the peak and end sentiments.

Based on the identified causal mechanism behind SA data from the causal discovery task, we further explore how it can improve prediction performance in the era of LLMs. Existing literature highlights ``causal alignment,'' namely to align the prediction direction along the underlying causal direction \cite{jin-etal-2021-causal,scholkopf2022causality,scholkopf2021towards}, but to our knowledge we are the first to explore how causal alignment improves model performance of SA in the era of LLMs. Specifically, we answer three subquestions: (Q1) If using the standard SA prompt, do models perform differently on C1/C2 data? (Q2) Does it help if we make the prompt aware of the underlying causality, i.e., use \textit{causal prompts}? And (Q3) When prompted causally, 
do LLMs mechanistically understand the corresponding causal processes?

Our empirical results show that under the standard prompt, LLMs perform better on data corresponding to the C2 causal process. Moreover, causal prompts aligned with the causal direction of the data can substantially improve the performance of zero-shot SA. Finally, we apply mechanistic interpretability methods to probe the models, and find that there is still improvement space for LLMs to correctly grasp the essence of 
the two causal processes. 
In summary, the contributions of this paper are as follows:
\begin{enumerate}
    \item 
    We propose the dual nature of SA as a combination of two tasks: a causal discovery task, and a prediction task.
    \item 
    For causal discovery, we ground the two possible causal processes in psychology, and use the peak-end rule to identify them.
    \item 
For the prediction task, we inspect existing LLMs' performance on data corresponding to the two underlying causal processes, and design \textit{causal prompts} to improve model performance by up to 32.13 F1 points.
\end{enumerate}

\section{Problem Formulation of SA}
In this section, we formulate SA as a combination of two tasks: the traditional prediction task in NLP and the causal discovery task in statistics, which we will introduce in the following.

\subsection{The Prediction Task (in NLP)}\label{sec:pred}
SA is a prediction task to identify the sentiment $y$ given a piece of text $x$.
We adopt the setup in most existing SA datasets \cite{maas2011learning,zhang2015character,keung-etal-2020-multilingual}, where the text $x$ is a review consisting of $n$ sentences $(t_1, \dots, t_n)$, and the label $y$ is a sentiment score corresponding to the star rating of the review in 1 (most negative), 2, \dots, 5 (most positive).

\subsection{The Cause-Effect Discovery Task (in Statistics)}\label{sec:discovery}

As a separate problem, there is an established task in causal discovery, the causal-effect problem \cite[see the review by][]{janzing2019causeeffect}, which aims to tell the cause from effect using only observational data. Its formal formulation is as follows:
Suppose we have an i.i.d. dataset $\mathcal{D} :=\{(x_i, y_i)\}_{i=1}^n$ containing $n$ observational data pairs of the two variables, $X$ and $Y$. The task is to \textit{infer whether $X$ causes $Y$ (i.e., $X \rightarrow Y$), or $Y$ causes $X$ (i.e., $Y \rightarrow X$)}, if one out of the two is true. In causality, ``$\rightarrow$'' indicates the directional causal relation between two variables. The two hypotheses can also be expressed in their equivalent structural causal models \cite[SCMs;][]{pearl2000causality} as introduced in \citet{peters2017elements}:
\begin{align}
&\text{Causal Hypothesis 1 (C1): } 
 X \rightarrow Y  
\\
& \Leftrightarrow 
Y: = f_Y(X, N_Y) \text{ with } N_Y \perp X~,
\\\nonumber
\\
&\text{Causal Hypothesis 2 (C2): } 
 Y \rightarrow X  
\\
& \Leftrightarrow 
X: = f_X(Y, N_X) \text{ with } N_X \perp Y~,
\end{align}
where $N_{i}$ is an unobserved noise term orthogonal to the input distribution.

\subsection{Causality and NLP Model Performance}\label{sec:causality_helps_nlp}

For many years, causality and machine learning have been two separate domains on their own.
Recently, researchers started to think about how the causal knowledge of the data can improve machine learning performance on the prediction task, especially for the two variable cause-effect case \cite{schoelkopf2012causal,jin-etal-2021-causal,ni-etal-2022-original}.
The essence of this line of research is that causality makes the two learning tasks $x \mapsto y$ and $y \mapsto x$ asymmetric, as one function's prediction direction \textit{aligns with} the ground-truth causal direction behind the two random variables, and another contradicts.
We call this phenomenon ``\textit{causal alignment},'' or ``direction match,'' of the prediction task and the causality.

To contrast the contribution of our work, we review the 
previous literature on causal alignment, which only shows its effect on the performance of trained-from-scratch machine learning models, without any indications in the era of LLMs:

\begin{enumerate}
    \item 
Causal alignment makes a model more robust against \textbf{covariant shifts} \cite[\textit{inter alia}]{jin-etal-2021-causal,scholkopf2022causality,schott2018towards} 
\item 
Semi-supervised learning (SSL) only works under causal misalignment, as the cause variable contains no information about the mechanism, but the effect variable does. So in the misaligned case, additional $P_X$ (i.e., the effect variable) helps \textbf{SSL} \cite{schoelkopf2012causal,jin-etal-2021-causal}.

\item 
Learning a causally-aligned model induces \textbf{less Kolmogorov complexity} (a more minimal description length) than the causally-misaligned model on the same $X$-$Y$ data \cite{jin-etal-2021-causal,janzing2010causal}
\item
Causal alignment significantly affects model performance in \textbf{supervised learning}, in the case of machine translation \cite{ni-etal-2022-original}.
\end{enumerate}

All the above findings are drawn under the training condition that we can isolate the training data to be only of one causal direction.
In the era of LLMs, we have seen substantial differences: (1) the training data can be a mixture of both causal directions, (2) the operationalization of the prediction task is through prompting, but no longer a separate model for each direction, and,  (3) in general, research has shifted to designing better prompts for already pre-trained models in their inference mode. 

Given these changes, we use the rest of the paper to address the following research questions:
\begin{enumerate}
    \item What is the causal direction in SA? (\cref{sec:psych})
    \item Can causal alignment help us improve SA prompts in the era of LLMs? (\cref{sec:improve})
\end{enumerate}

\section{Causal Discovery: Does Sentiment Cause the Review, or Vice Versa?}

\label{sec:psych}

\subsection{Problem Setup
}\label{sec:paradigm}
As mentioned previously, the setup of the bivariate causal discovery problem is to infer whether $X$ causes $Y$ (C1), or $Y$ causes $X$ (C2), based on a dataset $\mathcal{D} :=\{(x_i, y_i)\}_{i=1}^n$ containing only observational data of the joint distribution.

\paragraph{Challenges}
The common paradigm to check causal discovery results is to generate simulated data, of which the ground truth causal graph is known \cite{zhang2009causality,spirtes2016causal}.
However, in the context of the established SA datasets, such as Yelp \cite{zhang2015character}, Amazon \cite{keung-etal-2020-multilingual}, and App Review \cite{app_review}, we would not be able to track each individual user and survey their original causal process when composing the review and the rating. 
Another solution would also be difficult, as it would require SA to abandon all the above well-established datasets, and meticulously collect new data while surveying the users' underlying causal process.

\paragraph{Our Approach}
In the context of our work, we propose that there are still rich findings that we could derive from the observation-only data in the existing datasets, without interviewing or conducting new costly data collection.

The key to our approach is the psychology theories of the two causal processes, as the relation between sentiment and text has been well-studied and verified by randomized control trials (RCTs), among many other experiments.
In the rest of the section, we first introduce in \cref{sec:psych_theory} the psychology theories of fast and slow thinking, followed by the Peak-End Rule as the quantitative signal. Then, we operationalize the theory with computational techniques in \cref{sec:emotion_arc}, and the report findings on three different SA datasets in \cref{sec:psych_exp}.

\subsection{Psychological Processes Underlying Sentiment Processing}\label{sec:psych_theory}

\paragraph{Two Systems of Emotional Responses}
In psychology, the bifurcation into System 1 and System 2 in human decision-making, including sentiment processing, has garnered substantial empirical support \cite{kahneman2011thinking,epstein1994integration}. 

System 1, or the ``\textit{Fast Thinking}'' system, operates involuntarily, effortlessly, and without conscious awareness. It is often optimized in evolution to provide rapid responses to environmental stimuli \cite{ledoux1998emotional}, and guides most of our daily cognitive processing \cite{kahneman2011thinking}, and emotional responses such as fear or joy \cite{zajonc1980feeling}.

Conversely, System 2, often termed as ``\textit{Slow Thinking},'' is deliberate, slower, and more rational, requires more conscious effort \cite{kahneman2011thinking}, and allows for self-regulation and thoughtful consideration before making decisions \cite{baumeister1998ego}.
The interplay between these systems influences everything from mundane to critical decisions, highlighting the complexity of human emotional and cognitive processing \cite{kahneman2002representativeness,kahneman2011thinking}.

\paragraph{Correspondence to the Two Causal Processes}
There is a nice correspondence between the fast/slow thinking systems and our two causal hypotheses. As mentioned previously, the Causal Hypothesis 2 (C2) posits $Y \rightarrow X$, where the sentiment $Y$ causes the review $X$, which aligns well with the \textit{Fast Thinking} system \cite{kahneman2011thinking,ledoux1998emotional}, as it rapidly generates an emotional reaction $Y$, and then writes text to justify it $Y$. 
On the other hand, the Causal Hypothesis 1 (C1) refers to the case where $X \rightarrow Y$, namely the review $X$ causing the sentiment $Y$. It is an instance of \textit{Slow Thinking} \cite{baumeister1998ego,kahneman2002representativeness}, which deliberately uses conscious efforts to list out the up- and downsides of an experience in the review $X$, and come up with a thoughtful final decision as the rating $Y$.

\paragraph{Quantitative Signals of the Two Processes}
In sentiment processing, an evidence for the two processes is the famous \citet{kahneman1993more} study illustrating the \textit{Peak-End Rule} of
how individuals recall and evaluate past
emotional experiences, which we show in \cref{fig:structure}. 
As we know, fast thinking is prone to systematic biases and errors in the judgment \cite{tversky1974judgment}, and the \citet{kahneman1993more} study provides important quantitative results showing that,
in the Fast Thinking system, people's emotional memories of an experience are disproportionately influenced by its most intense point (the ``peak'') and its conclusion (the ``end''), rather than by the average experience as in the Slow Thinking system. The important role of peak and end for the fast thinking system implies that
it is the intensity of specific moments that dominate memory and judgment.

\begin{table*}[t]
    \centering \small
    \setlength\tabcolsep{2.4pt}
    \begin{tabular}{lccccccccc}
\toprule

& \multicolumn{3}{c}{Yelp} & \multicolumn{3}{c}{Amazon} & \multicolumn{3}{c}{App Review}
\\
\cmidrule(lr){2-4} \cmidrule(lr){5-7} \cmidrule(lr){8-10}
& All & C1 & C2 & All & C1 & C2 & All & C1 & C2 \\
\midrule
\# Samples & 34,851 & 19,557 (56\%) & 15,294 (44\%) & 2,582 & 1,393 (54\%) & 1,189 (46\%) & 9,696 & 3,809 (39\%) & 5,887 (61\%)\\
\# Sents/Review & 11.11  & 11.30 & 10.87 & 6.70 & 6.62 & 6.80 & 6.34 & 6.33 & 6.35\\ 
\# Words/Sent & 15.53 & 15.55  & 15.49 & 11.04 & 11.28 & 10.77 & 10.53 & 10.90 & 10.29\\
Vocab Size & 64,864 & 48,889 & 44,826 & 10,271  & 7,609  & 7,049 & 20,248 & 12,773 & 15,400\\
Avg Sentiment & 2.93 & 2.74 & 3.18 & 2.94 & 2.82 & 3.07 & 2.9 & 2.72 & 3.02\\ \hline
Avg $\lambda_1$ & 3.78 & 2.97 &  4.83 & 3.77 & 3.10 & 4.55 &  6.03 & 5.18 & 6.58 \\
Avg $\lambda_2$ &  4.48 & 6.05 & 2.48 & 4.21 & 5.72 & 2.45 & 5.10 & 7.96 & 3.26\\

\bottomrule
    \end{tabular}
    \caption{Statistics of the entire datasets and their C1 and C2 subsets for Yelp, Amazon, and App Review. We can see that a roughly balanced number of reviews aligning with the C1 and C2 processes.
    }
    \label{tab:psych}
\end{table*}
\subsection{Operationalizing the Theory}

We summarize the previous psychological insights in the upper left part of \cref{fig:structure}, where the Causal Hypothesis 1 corresponds to taking the average of all emotional experiences mentioned in the review $X$ for the sentiment $Y$, and the Causal Hypothesis 2 uses the peak and end emotions in the review $X$ to derive the sentiment $Y$.
In this section, we introduce a formalization of the theory, and suggest signals to distinguish the two causal hypotheses.

\paragraph{Emotion Arc}\label{sec:emotion_arc}
To capture the aforementioned trajectory of emotional experiences, we use the concept of the \textit{emotion arc} \cite{reagan2016emotional}, an example of which we visualize in \cref{fig:structure}.
Contextualizing it in the task of SA, we formally define an emotion arc of the review as follows.
Given a review $x$ consisting of $n$ sentences $(t_1, \dots, t_n)$, we identify the sentiment for each of them, thus obtaining a series of sentiment labels $(s_1, \dots, s_n)$.
We denote this series as the emotion arc $\bm{e}:=(s_1, \dots, s_n)$ of the review.

\paragraph{The Two Causal Processes}
Provided the notion of the emotion arc $\bm{e}:=(s_1, \dots, s_n)$ for a review $x$,
we formulate the sentiment labels corresponding to the two causal processes as follows:
\begin{align}
 \nonumber
    \text{Slow Thinki} &\text{ng (Causal Process 1): }
    \\
    \hat{y}_{\mathrm{avg}} &= \frac{1}{n} \left( s_1+ \dots+s_n \right)
    ~,
    \\
    \lambda_1 &= |y - \hat{y}_{\mathrm{avg}}|
    ~,
    \\
\nonumber
    \text{Fast Thinki} &\text{ng (Causal Process 2): }
    \\
\hat{y}_{\mathrm{peakEnd}} &= \frac{1}{2} \left( \mathrm{Peak} (s_1, \dots,s_n) + s_n \right)
    ,\\
    \lambda_2 &= |y - \hat{y}_{\mathrm{peakEnd}}|
    ~,
    \end{align}
where $\lambda_i$ indicates the alignment of the actual sentiment $y$ with the Causal Process $i$, and $\mathrm{Peak}(\cdot)$ 
selects the sentiment with the strongest intensity by its distance from the neutral sentiment 3, which is the middle point among the sentiment range 1--5, i.e.,
$\mathrm{Peak} (s_1, \dots,s_n) := 
s_{\argmax_i |s_i - 3|}
$.

\begin{table}[t]
    \centering \small
    \begin{tabular}{p{7.5cm}lllllll}
\toprule
\multicolumn{1}{c}{\textbf{Example of a C1-Dominant Review}}
\\
$\bm{\cdot}$ This was a great spot to take a break from it all and just people watch. $s_1=4.57$
\\
$\bm{\cdot}$ We sat at the bar facing the casino and we were entertained the whole time.$s_2=4.67$\\
$\bm{\cdot}$ The mini grilled cheese (appetizer) was fantastic. $s_3 = 4.53$\\
$\bm{\cdot}$ It came with a tomato based dipping sauce that was the perfect compliment to the bite sized wedges. $s_4 = 4.20$\\
$\bm{\cdot}$ Tip - ask for two dipping sauces because one just won't do. $s_5 = 1.60$\\

\textbf{Stars $y$:} 4 \\
\textbf{Psychology Scores ($\downarrow$):} $\lambda_1 = 0.0884  < \lambda_2 = 0.8683$ \\
\midrule
\multicolumn{1}{c}{\textbf{Example of a C2-Dominant Review}}
\\
$\bm{\cdot}$ I read the reviews and should have steered away... but it looked interesting. $s_1 = 3.72$
 \\
$\bm{\cdot}$ Salad was wilted, menus are on the wall, with no explanation so you are ordering blind, service was NOT with a smile from the bartender to the waitress, to the server who helped the waitress, and the waitress never checked back to see how everything is. $s_2 = 2.20$\\
$\bm{\cdot}$ Terribly overpriced for what you get, and as an Italian, this does not even pass for a facsimile thereof! $s_3 = 1.45$\\
$\bm{\cdot}$ Stay away for sure. $s_4 = 1.85$\\
$\bm{\cdot}$ I only gave them one star, as I had to fill something in, they should get no stars! $s_5 = 1.32$\\

\textbf{Stars $y$:} 1 \\
\textbf{Psychology Scores ($\downarrow$):} $\lambda_1 = 1.1827 > \lambda_2 = 0.3647$ \\
\bottomrule
    \end{tabular}
    \caption{Examples of C1- and C2-dominant reviews.}
    \label{tab:psych_examples}
\end{table}

Here, we interpret $\lambda_i$ as an indicator for each causal process, where a small value (with the best value being zero) implies the alignment with the process $i$.
We show two examples in \cref{tab:psych_examples}, one aligning well with the Causal Process C1 with a small $\lambda_1$, and another aligning well with the Causal Process C2 with a small $\lambda_2$.

\begin{table*}[t]
    \centering \small
    \begin{tabular}{llccccccccc}
    \toprule
 && Random & GPT-2 XL & LLaMa-7B & Alpaca-7B & GPT-3 & GPT-3.5 & GPT-4\\ \midrule
\multirow{3}{*}{F1} & Overall & 19.82 {\tiny$\pm$2.07} & 10.23 {\tiny$\pm$4.12} & 31.78 {\tiny$\pm$5.32} & 46.01 {\tiny$\pm$5.35} & 52.71 {\tiny$\pm$1.73} & 57.98 {\tiny$\pm$5.11} & 59.54 {\tiny$\pm$4.69}
    \\
& C1 Subset  & 21.36 {\tiny$\pm$2.26} & 5.80 {\tiny$\pm$3.11} & 27.30 {\tiny$\pm$4.73} & 37.77 {\tiny$\pm$7.66} & 43.96 {\tiny$\pm$2.93} & 58.64 {\tiny$\pm$1.48} & 58.62 {\tiny$\pm$2.54}  \\
& C2 Subset &  20.43 {\tiny$\pm$2.95} & \textbf{16.37} {\tiny$\pm$5.33} & \textbf{37.66} {\tiny$\pm$7.86} & \textbf{55.82} {\tiny$\pm$4.02} & \textbf{65.40} {\tiny$\pm$1.37} & \textbf{59.09} {\tiny$\pm$9.13} & \textbf{62.57} {\tiny$\pm$6.85} \\ \hline
\multirow{3}{*}{Accuracy} & Overall &19.78 {\tiny$\pm$2.07} & 23.06 {\tiny$\pm$2.10} & 39.28 {\tiny$\pm$5.07} & 47.72 {\tiny$\pm$4.19} & 53.22 {\tiny$\pm$1.35} & 58.36 {\tiny$\pm$4.13} & 59.84 {\tiny$\pm$4.17}
\\
& C1 Subset & 20.61 {\tiny$\pm$2.23} & 16.18 {\tiny$\pm$1.59} & 36.55 {\tiny$\pm$4.05} & 42.14 {\tiny$\pm$5.26} & 43.61 {\tiny$\pm$2.89} & \textbf{59.89} {\tiny$\pm$1.09} & 59.62 {\tiny$\pm$1.96} \\
& C2 Subset &  18.86 {\tiny$\pm$2.78} & \textbf{30.79} {\tiny$\pm$2.68} & \textbf{42.33} {\tiny$\pm$7.24} & \textbf{53.93} {\tiny$\pm$3.91} & \textbf{63.93} {\tiny$\pm$1.28} & 56.66 {\tiny$\pm$8.08} & \textbf{60.08} {\tiny$\pm$7.05}\\

    \bottomrule
    \end{tabular}
    \caption{Performance of different models on the five-class classification of Yelp-5. We use five paraphrases for the prompt (in \cref{appd:prompt_we_use}), and report the average performance with the standard deviation.
    }
    \label{tab:prompt_c0}
\end{table*}
\begin{table*}[t]
    \centering \small
    \setlength\tabcolsep{4.8pt}
    \begin{tabular}{llcccccccccccc}
\toprule
&&Random & GPT-2 XL & LLaMa-7B & Alpaca-7B & GPT-3 & GPT-3.5 & GPT-4 \\ \midrule
\multirow{4}{*}{F1} 
& Data=C1, Prompt=C1 &20.47 {\tiny$\pm$2.47} & 6.12 {\tiny$\pm$2.77} & 55.16 {\tiny$\pm$7.16} & 52.74 {\tiny$\pm$5.04} & 38.36 {\tiny$\pm$6.66} & 60.62 {\tiny$\pm$3.24} & 54.23 {\tiny$\pm$4.17} \\
& Data=C1, Prompt=C2 &20.26 {\tiny$\pm$2.31} & 15.98 {\tiny$\pm$3.59} & 36.22 {\tiny$\pm$8.95} & 35.10 {\tiny$\pm$5.40} & 54.44 {\tiny$\pm$1.64} & 52.85 {\tiny$\pm$6.01} & 58.58 {\tiny$\pm$3.33}\\
& Data=C2, Prompt=C1 & 22.35 {\tiny$\pm$3.02} & 31.98 {\tiny$\pm$8.66} & 56.69 {\tiny$\pm$8.46} & 54.74 {\tiny$\pm$14.26} & 74.64 {\tiny$\pm$3.06} & \textbf{78.18} {\tiny$\pm$1.21} & 72.52 {\tiny$\pm$3.68} \\
& Data=C2, Prompt=C2 & 20.35 {\tiny$\pm$2.18} & \textbf{48.50} {\tiny$\pm$7.66} & \textbf{66.82} {\tiny$\pm$7.90} & \textbf{71.22} {\tiny$\pm$3.99} & \textbf{77.09} {\tiny$\pm$1.33} & 78.16 {\tiny$\pm$1.81} & \textbf{76.80} {\tiny$\pm$1.36}\\ \hline

\multirow{4}{*}{Acc}
& Data=C1, Prompt=C1 &19.60 {\tiny$\pm$2.67} & 12.96 {\tiny$\pm$2.91} & 58.23 {\tiny$\pm$5.30} & 55.81 {\tiny$\pm$4.00} & 43.16 {\tiny$\pm$6.20} & 60.39 {\tiny$\pm$3.25} & 54.06 {\tiny$\pm$3.97}  \\
& Data=C1, Prompt=C2 & 19.68 {\tiny$\pm$2.46} & 22.61 {\tiny$\pm$5.73} & 43.07 {\tiny$\pm$8.18} & 41.45 {\tiny$\pm$4.06} & 56.36 {\tiny$\pm$1.94} & 53.11 {\tiny$\pm$5.89} & 58.68 {\tiny$\pm$3.45}\\
& Data=C2, Prompt=C1 & 20.97 {\tiny$\pm$3.19} & 43.51 {\tiny$\pm$6.30} & 57.54 {\tiny$\pm$7.21} & 54.82 {\tiny$\pm$12.59} & 76.60 {\tiny$\pm$3.23} & 77.38 {\tiny$\pm$1.43} & 70.69 {\tiny$\pm$3.93}\\
& Data=C2, Prompt=C2 & 19.03 {\tiny$\pm$2.18}& \textbf{51.59} {\tiny$\pm$5.09} & \textbf{68.16} {\tiny$\pm$9.24} & \textbf{71.83} {\tiny$\pm$4.33} & \textbf{76.70} {\tiny$\pm$1.35} & \textbf{78.92} {\tiny$\pm$1.31} & \textbf{75.38} {\tiny$\pm$1.70} \\
\bottomrule
    \end{tabular}
    \caption{
    Performance on Yelp using the \textit{causal prompts} on the two causal subsets. We report the average performance across the five paraphrases for each prompt, with the standard deviation.
    }
    \label{tab:prompt_c12}
\end{table*}

\subsection{Findings on SA Datasets}\label{sec:psych_exp}
\paragraph{Dataset Setup}
We adopt three commonly used datasets in SA: Yelp \cite{zhang2015character}, Amazon \cite{keung-etal-2020-multilingual}, and App Review \cite{app_review}.
For the Amazon data, we concatenate each review's title with its text.
Since the model performance on many binary classification datasets is saturated \cite{poria2023beneath,yang2019xlnet}, we use the 5-way classification version of the SA datasets when applicable.

Since we need to utilize the emotion arc, we keep only reviews with at least five sentences, after sentence tokenization using the Spacy package \cite{spacy2}.
We apply this filtering above on the test set of the Yelp dataset, the English test of Amazon, and the unsplit entire dataset of App Review. We report the statistics of remaining samples in \cref{tab:psych}.

To obtain the emotion arcs, we calculate the sentiment score of each sentence produced by the \texttt{sentiment-analysis} pipeline\footnote{\href{https://huggingface.co/distilbert-base-uncased-finetuned-sst-2-english}{https://huggingface.co/distilbert-base-uncased-finetuned-sst-2-english}. See details in \cref{appd:sent_score}.} from Huggingface \cite{wolf-etal-2020-transformers}.

\paragraph{Causal Discovery}
For each input sample, we process them as in \cref{tab:psych_examples}, namely first obtaining the sentence-level sentiments to form the emotion arc, and then calculating the alignment scores $\lambda_1$ and $\lambda_2$ for each causal process, respectively.
We consider an example as \textit{dominated} by the causal process $C_i$ if the alignment score $\lambda_i$ is more optimal than the other. We report the resulting statistics in \cref{tab:psych}. For each dataset, we describe their overall statistics, as well as the statistics of data with the underlying causal process of C1, and that of C2. We can see that Yelp and Amazon have an almost balanced split of C1 and C2, while App Review has 61\% C2 data compared to 39\% C1 data. See \cref{appd:distr_plot} for an additional visualization of the $\lambda_1$-$\lambda_2$ distribution across the 1K data points.

\section{How to Improve Sentiment Classifiers with Causal Alignment?}
\label{sec:improve}
Using our proposed causal discovery method, we have identified two distinct subsets with their corresponding causal processes C1 and C2.
Now, we address the last practical question proposed in \cref{sec:causality_helps_nlp}:
\begin{quote}
    \textit{Can causal alignment help us improve SA in the era of LLMs?}
\end{quote}

Specifically, we take the commonly used approach in the era of LLMs, i.e., prompting pre-trained LLMs for the SA task, and look into how alignment with the underlying causal process could help SA performance. We answer the following three subquestions in this section:
\begin{enumerate}[label=Q\arabic*.]
    \item 
    Using the standard SA prompt, do models perform differently on C1/C2 data? (\cref{sec:q1})
    \item 
    Does it help if we make the prompt aware of the underlying causality, i.e., use causal prompts? (\cref{sec:q2})
    \item 
    When prompted causally, 
    do LLMs really understand the causal processes? (\cref{sec:q3})
\end{enumerate}

\subsection{Q1: Do Models Perform Differently on C1/C2 Data?}\label{sec:q1}

\paragraph{Experimental Setup}
The first question is whether models perform differently on data with the causal nature of C1 or C2. We use the subsets identified by our psychologically-grounded causal discovery, and test a variety of available autoregressive LLMs, including the open-weight GPT-2 \cite{radford2019language},
LLaMa \cite{touvron2023llama}, and
Alpaca \cite{alpaca}; as well as the closed-weight models with OpenAI API, the instruction-tuned GPT-3 ({text-davinci-002}) \cite{brown2020GPT3,ouyang2022instructGPT}, GPT-3.5 ({gpt-3.5-turbo-0613}), GPT-4 ({gpt-4-0613}) \cite{openai2023GPT4}.
We also add a random baseline which uniformly samples the label space for each input.

We use the standard prompt formulation for SA in the format of ``\texttt{[Instruction] Review Text: \{$x$\}$\backslash$n Label:}''. The experiments are on a set of randomly selected 1K samples from the test set of Yelp-5 \cite{zhang2015character}, due to the time- and cost-expensive inference of the above LLMs.
(E.g., LLaMa/Alpaca takes 96 GPU hours to run.)
See more experimental details in
\cref{appd:models}.

\paragraph{Results}
We show the performance of the six LLMs in \cref{tab:prompt_c0}, and report the F1 and accuracy across the five-class classification on Yelp-5.
We can see that the existing LLMs perform the best on the subset with the causal process C2, implying that the decision pattern of LLMs is closer to the Fast Thinking system, which takes the peak-end average of the emotion arc.

\subsection{Q2: Do Causal Prompts Help?}\label{sec:q2}

\paragraph{Designing \textit{Causal Prompts}}
Inspired by the fact that models perform differently on C1/C2 data, our next question is, will it help if we directly give a hint to the LLMs about the underlying causal graph?

\begin{table}[ht]
    \centering \small
    \begin{tabular}{m{0.6cm}p{6cm}}
\toprule
& Prompt Design \\ \midrule
    C1     & 
As a customer writing a review, I initially \textit{composed} the following feedback: ``\texttt{[review]}'' \vspace{1mm}
\newline
\textit{After carefully considering the facts}, I selected a star rating from the options of ``1'', ``2'', ``3'', ``4'', or ``5''. My final rating was:
\\\hline
    C2 & 
As a customer writing a review, I initially \textit{selected} a star rating from the options ``1'', ``2'', ``3'', ``4'', and ``5'', and then provided the following explanations in my review: ``\texttt{[review]}'' 
\vspace{1mm}
\newline
The review \textit{clarifies} why I gave a rating of
\\
\bottomrule
    \end{tabular}
    \caption{Causally-aware prompts describing the SA task in contexts with the C1 and C2 causal graphs.}
    \label{tab:prompt_c12_design}
\end{table}

To this end, we propose the idea of \textit{causal prompts}, which are prompts that describe the causal story behind the input and output variables. We list our designed prompts for the C1 and C2 stories in \cref{tab:prompt_c12_design}. 

\paragraph{Results}
We report the performance for all combinations of the dataset natures and prompt natures in \cref{tab:prompt_c12}, where we find that the most-performant setting uses Prompt C2 on the data subset with the same causal nature, C2. This alignment leads to the best performance across almost all models by both F1 and accuracy. 
On the C2 data, we also see that Prompt C2 outperforms the standard SA prompt in \cref{tab:prompt_c0} by a substantial margin, such as 32.13 F1 points increase for GPT-2, and 14.23 F1 points increase for GPT-4.

However, although Prompt C2 shows a strong performance, the other causal prompt, i.e., Prompt C1, does not always help the data subset C1 in all cases, from which we raise a further question -- how well do LLMs really mechanistically understand our prompts? We explore this question in the next section.

\subsection{Q3: Can LLMs Correctly Capture the Causal Stories in the Prompts?}\label{sec:q3}

Although the proposal of the two causal prompts is intuitive for humans, we still need to inspect whether LLMs are able to understand them correctly.

\paragraph{Method}
Mechanistically, for a model to solve SA for the causal process C1 correctly, it needs to treat the sentence-level sentiments across all sentences \textit{equally}; and for a model to solve SA for the causal process C2 correctly, it needs to pay \textit{more} attention to the peak and end sentiments on the emotion arc.

Targeting the two mechanisms, we use causal tracing \cite{meng2022locating} to attribute the final sentiment prediction to the source sentences in the input. Briefly, causal tracing uses causal mediation analysis \cite{pearl2001direct} to quantify the causal contribution of the internal neuron activations of a model to its final prediction \cite{vig2020causal}. 
We use causal tracing to inspect the causal effects of the hidden states on the model prediction, using the open-weight models, LLaMa and Alpaca. We use the causal effects of the first-layer neurons for each sentence, which we aggregate to obtain the final prediction.
See implementation details in \cref{appd:causal_tracing}.

\begin{figure}[t]
    \centering
    \includegraphics[width=0.95\columnwidth]{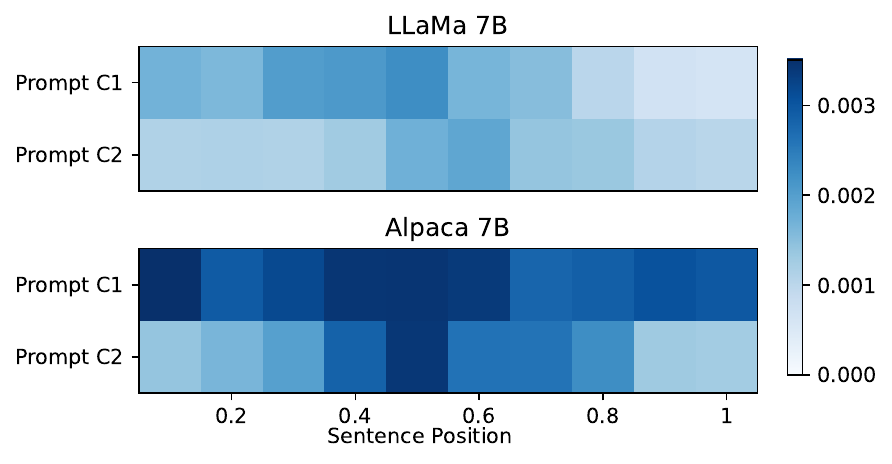}
    \caption{Causal attribution in LLaMa-7B and Alpaca-7B, showing how much each sentence contributes to the prediction probability. 
}
    \label{fig:causal_trace}
\end{figure}

\paragraph{Results}
We plot the causal attribution results of how much each sentence contributes to the final prediction
in \cref{fig:causal_trace}.
Here, the ideal behavior of the models is that Prompt C1 should trigger \textit{uniform} attention over all the sentences, which is roughly observed through the more even shades of color of the ``Prompt C1'' row than the ``Prompt C2'' row in \cref{fig:causal_trace} in the row of ``Prompt C1''. 

On the other hand, Prompt C2 should trigger \textit{more} attention to the sentences corresponding to the peak and end sentiments. For this, we see the models have high attention to the middle sentence, as in the ``Prompt C2'' row in \cref{fig:causal_trace}. 
The average causal effect of the peak sentence on predictions under Prompt C2 is 3\% higher than the mean effect for the LLaMa model, and 39\% higher for the Alpaca model.
This aligns with our expectation that the peak sentence would have a high contribution.
Nonetheless, note that no model sufficiently attends to the end sentence under Prompt C2. This implies that they do not fully grasp the expected contribution pattern of the peak-end rule, missing the significant role of the end sentence.

\section{Related Work}

\paragraph{SA}
The task of SA aims to identify the sentiment given a piece of text.
It has a rich history originating from subjectivity analysis \cite{wiebe1994tracking,hatzivassiloglou2000effects}, and developing rapidly with the availability of large opinionated online data such as reviews with star ratings \cite[\textit{inter alia}]{turney2002thumbs,nasukawa2003sentiment,zhang2015character,keung-etal-2020-multilingual}. Most literature on SA focuses on building computational models, from using traditional linguistic rules \cite{hatzivassiloglou1997predicting,DBLP:conf/emnlp/ChoiC08}, to the application of machine learning methods, from traditional naive bayes and support vector machines \cite{pang2002thumbs,moraes2013document,DBLP:conf/ecir/TanCWX09}, to early deep learning models
\cite{socher2013recursive,kim2014convolutional,xing-etal-2020-tasty}, and finally entering the era of LLMs \cite{DBLP:conf/nodalida/HoangBR19,raffel2020exploring,yang2019xlnet}.

\paragraph{Psychology and Affective Science}
In the study of emotion, or affect science \cite{salovey2004emotional,barrett2006solving,feinstein2013lesion},
previous work
finds that not only the emotion people perceive influences or prime 
how they communicate in the
moment \cite{barrett2006solving}, but language can also influence
emotion, which can be observed in functional magnetic neuroimaging \cite{satpute2013functional}, and also experiments showing the act of self-reporting the emotion in writing can change the physical reaction to the emotion \cite{kassam2013effects}.
In his seminal work, \citet{kahneman2011thinking} uses the two systems of thinking to reveal the mechanisms of how people come up with their sentiment, where fast thinking conforms to the peak-end rule \cite{kahneman1993more}, and slow thinking is more reflective of the overall sentiment.

\paragraph{Cause-Effect Distinction}
Distinguishing the cause from effect based on observational data is a long-standing and fundamental problem in causality \cite{hoyer2008nonlinear,zhang2009causality,janzing2019causeeffect}. Existing methods to address this problem are based on statistics \cite{hoyer2008nonlinear,peters2010identifying,shajarisales2015telling,mooij2014distinguishing}, physics \cite{janzing2007causally,janzing2016algorithmic}, information theory \cite{janzing2012informationgeometric,chaves2014inferring,mejia2022obtaining}, and algorithmic complexity \cite{janzing2010causal,jin-etal-2021-causal}. 
However, we are the first to look at the rich nature of NLP datasets, and directly approach the difference in the causal and anticausal mechanisms grounded in interdisciplinary insights. 

As for our causal prompts, the most similar studies are the non-causally-grounded explorations for prompt tuning, such as by varying the patterns of masked language modeling \cite{schick-schutze-2022-true} and using the noisy channel method \cite{min-etal-2022-noisy}. However, these studies are not aware of the underlying causal processes, thus neglecting the connection of prompts with the causal nature of data, and also the explicit causal story of the sentiment-review relation.

\section{Conclusion}
In conclusion, we have formulated the task of SA into a prediction problem and a causal discovery problem. We first identified the cause-effect relation among existing SA datasets, namely 
whether the review primes the rating, or the sentimental judgment primed the review writing process.
To achieve this causal discovery, we obtain insights from existing psychology studies, namely aligning the above two causal processes with the famous Fast Thinking and Slow Thinking systems, with their distinct qualitative signals.
Given the causal understanding of the dataset, we further improve the performance of LLMs on SA using our proposed causal prompts. Our research paves the way for more causally-aware future research in SA.

\section*{Limitations and Future Work}

This study has several limitations. First, the rapid progression of LLMs makes it challenging to keep up with all newly proposed models and architectures. Since our work covers only a set of recent LLMs at the time of this study, we encourage future research to apply our methods to additional LLMs and other SA datasets.

Although our study is grounded in well-established psychological theories, there remains the possibility that new theories could emerge, necessitating updates to the calculation of the $\lambda$ values for the two causal processes. However, the causal processes identified in this work appear plausible, as evidenced by the effectiveness of the causally aligned prompts in improving language model performance.

Regarding the causal graph formulation, we focus on basic bivariate causal graphs, but future work could include more variables, such as confounders, mediators, and colliders. 

The nature of this work is to introduce a paradigm shift for SA, and formulate the task differently. 
Therefore, we see lots of space for future extensions, such as to explore the causal nature of SA in different settings, different languages, and also aspect-based sentiment analysis \cite{DBLP:conf/semeval/PontikiGPPAM14,xing-etal-2020-tasty,hua2023systematic}.

\section*{Ethical Considerations}
Regarding data concerns and user privacy, our study employs several established NLP datasets, and the examples we cite do not include sensitive user information.

Concerning potential stakeholders and misuse, this research primarily introduces a new perspective on the SA task. A possible negative impact concerns the general application of SA, which could be used to analyze user mentality for surveillance or fraudulent purposes. We acknowledge that studies on SA inherently involve these risks, and we firmly oppose the misuse of SA models in such contexts.

\ifarxiv

\section*{Acknowledgment}
We thank Luigi Gresele and Dominik Janzing for constructive discussions of the formulation and landscape of existing literature of causality.
We thank Julius von Kügelgen for brainstorming the idea of quantifiable footprints for causal and anticausal relations as a followup of \citet{jin-etal-2021-causal}.
The work also benefits from the discussion with Jacob Eisenstein on whether NLP datasets are causal or anticausal as a followup of \citet{veitch2021counterfactual}.
For the domain knowledge on psychology, especially affect science, 
we thank Prof Erik C. Nook at the Department of Psychology at Princeton University for the discussions and pointers.
We thank labmates at LIT lab at University of Michigan for many helpful feedback on the writing, especially Joan Nwatu, Siyang Liu, and Aylin Gunal.

This material is based in part upon works supported by the German Federal Ministry of Education and Research (BMBF): Tübingen AI Center, FKZ: 01IS18039B; by the Machine Learning Cluster of Excellence, EXC number 2064/1 – Project number 390727645; 
by the Precision Health Initiative at the University of Michigan; 
by the John Templeton Foundation (Grant \#61156); by a Responsible AI grant by the Haslerstiftung; and an ETH Grant
(ETH-19 21-1).
Zhijing Jin is supported by PhD fellowships from the Future of Life Institute and Open Philanthropy, travel support from ELISE (GA no 951847) for the ELLIS program, and API credits by the OpenAI Researcher Access Program.

\section*{Author Contributions}\label{sec:contributions}
For the idea formulation, Zhijing Jin drew inspirations from her previous work \cite{jin-etal-2021-causal} and proposed to extend it to LLMs.
During the internship of Zhiheng Lyu at ETH Zürich hosted under Prof Mrinmaya Sachan and mentored by Zhijing, Zhiheng first explored whether LLMs show a distinct fingerprint for causal and anticausal prompts in a workshop paper \cite{lyu2022can}. Then Zhijing got inspired by chats with psychology researchers in affect science that the phenomena of sentiment-primed text and text-primed sentiments can be grounded in actual psychology studies. Hence, we started to explore this paper's idea together.

\textit{Zhiheng Lyu} conducted all the experiments and analyses, proposed various novel experimental designs, structured the storyline, and did tremendous hard work to make the paper happen. 

\textit{Zhijing Jin} closely mentored the project, proposed the initial idea as well as the Fast and Slow Thinking formulation, and wrote the current version of the paper.

\textit{Fernando Gonzalez} joined the project in the final but crucial iteration, and helped with the implementation of the modeling part, mechanistic interpretability, as well as part of the data analysis.

Professors \textit{Mrinmaya Sachan}, \textit{Rada Mihalcea}, and  \textit{Bernhard Schölkopf} supervised the project and gave lots of constructive advice on the writing.

\fi

\bibliography{sec/refs_acl,sec/refs_this_paper,sec/refs_causality,sec/refs_zhijing,sec/refs_cogsci,sec/refs_sa,sec/refs_discourse}

\clearpage
\appendix

\section{Implementation Details}

\subsection{Model Details}\label{appd:models}
\paragraph{Using Closed-Weight Models}
For the use of GPT model series, we use the OpenAI API,\footnote{ \url{https://openai.com/api/}} with a text generation temperature of 0.
We spent around 400 USD across around 20-30K single API calls.

\paragraph{Using Open-Weight Models}\label{appd:exp_details}
For reproducibility, we set the generation temperature to 0 for all the models used in our work. For the open-weight models, GPT2-XL, LLaMa-7B and Alpaca-7B, it took around 24 hours on 4 GPUs RTX 2080 to generate their predictions on 1K data points for the 5 paraphrases of the causally-neutral prompt (denoted as C0), and on 500 data points for the 5 paraphrases of the C1 prompt, and 5 paraphrases of the C2 prompt.
The causal tracing experiments with LLaMa-7B and Alpaca-7B on 100 data points took around 24 hours each using one GPU V100.

\subsection{Sentence Sentiment Score Calculation}\label{appd:sent_score}
We apply the inverse sigmoid function to convert sentiment score probabilities from the range [0,1] to \([- \infty, \infty]\). 
\begin{align}
\mathrm{logit}(p) = \log \left( \frac{p}{1 - p} \right)
~,
\end{align}
where \(p\) is the probability of the positive class.

In practice, we 
cap the output of the logit function to the range \([-10, 10]\) as the sentence sentiment score. Namely,
\begin{align}
\mathrm{score}(p) = \max(-10, \min(10, \log \left(\text{logit}(p) \right)))~.
\end{align}
In this way, our scores fall between -10 and 10, corresponding approximately to probabilities of the positive label between 0.0001 and 0.9999. 

To map our 5-class scores to the \([-10, 10]\) range, we assign the class labels as follows: -10 corresponds to the label 1, -5 to the label 2, 0 to the label 3, 5 to the label 4, and 10 to the label 5.

\subsection{Implementation Details for Causal Tracing}
\label{appd:causal_tracing}

We introduce the workings of the causal tracing method  \cite{meng2022locating} as follows.
First, we compute the hidden states of the residual stream of LLaMa-7B's layers for two inputs, (1) the original input: the prompt+review, and (2) the corrupted input: prompt + a corrupted version of the review by adding random noise immediately after the token embeddings. Then, we restore one by one the clean state of the residual stream into the corrupted version and measure the effect of the clean state on the probability of the originally predicted token for each token sequence and layer position. 

Since this process is highly time-consuming, taking around 12 hours for 50 samples even using the smallest LLaMa model with 7B parameters, we do a case study on the 7B LLaMa and Alpaca using 100 random samples from the 1K test set. 
For these experiments, we follow the idea of APE \cite{zhou2023large} to use the best-performing prompts on the 1k test set for C1 and C2.
\subsection{Prompts}\label{appd:prompt_we_use}

\subsubsection{Prompts to Get Paraphrases}
Since we need to report the average performance across five paraphrases of the same prompt, for each original prompt, we call GPT to generate the four paraphrases.

Below is the prompt that we used for this paraphrase generation process:
\begin{quote}
You are an expert in prompt engineering for large language models (LLMs). And you are also a native English speaker who writes fluent and grammatically correct text.

Given the following prompt for NLP sentiment analysis, you provide four alternative prompts.

\#\#\#\#\#\#\# Original Prompt \#\#\#\#\#\#\#
\texttt{[Our original prompt]}

\#\#\#\#\#\#\# Alternative Prompt 1 \#\#\#\#\#\#\#

... (Then, we let the model to generate all the way to ``Alternative Prompt 4''.)

\end{quote}

We queried the GPT-4 model with temperature 0\ifarxiv ~on June 8, 2023\fi.

\subsubsection{Neutral Prompt}
In addition to the standard prompt to query LLMs in the main paper, we show its four paraphrases in \cref{tab:prompts_paraphrases_c0}.

\begin{table}[ht]
    \centering \small
    \begin{tabular}{p{7cm}}
\toprule
Prompt Design \\ \midrule 
As a proficient data annotator in natural language processing (NLP), your responsibility is to determine the sentiment of the given review text. Please assign a sentiment value from ``1'' (very negative) to ``5'' (very positive).\newline Review Text: ``\texttt{[review]}'' \newline
Sentiment Score:
\\\hline
As a skilled data annotator in the field of natural language processing (NLP), your task is to evaluate the sentiment of the given review text. Please classify the sentiment using a scale from ``1'' (highly negative) to ``5''  (highly positive).\newline Review Text: ``\texttt{[review]}'' \newline
Sentiment Rating:
\\\hline
As an expert data annotator for NLP tasks, you are required to assess the sentiment of the provided review text. Kindly rate the sentiment on a scale of ``1'' (extremely negative) to ``5'' (extremely positive).\newline Review Text: ``\texttt{[review]}'' \newline
Sentiment Score:
\\\hline
As a proficient data annotator in natural language processing (NLP), your responsibility is to determine the sentiment of the given review text. Please assign a sentiment value from ``1'' (very negative) to ``5'' (very positive).\newline Review Text: ``\texttt{[review]}'' \newline
Sentiment Assesment:
\\
\bottomrule
    \end{tabular}
    \caption{Four additional paraphrases of the neutral prompt (C0) generated with GPT-4.}
    \label{tab:prompts_paraphrases_c0}
\end{table}

\subsubsection{Causal Prompts}
In addition to the standard C1 and C2 prompts in the main paper, we show the four paraphrases for each of them in  \cref{tab:prompts_paraphrases_c1,tab:prompts_paraphrases_c2}, respectively.

\begin{table}[ht]
    \centering \small
    \begin{tabular}{p{7cm}}
\toprule
Prompt Design \\ \midrule 
As a customer sharing my experience, I crafted the following review: ``\texttt{[review]}'' \newline
Taking into account the details of my experience, I chose a star rating from the available options of ``1'',``2'', ``3'', ``4'', or ``5''. My ultimate rating is:
\\\hline
As a client providing my opinion, I penned down the subsequent evaluation: ``\texttt{[review]}'' \newline
Upon thorough reflection of my encounter, I picked a star rating among the choices of ``1'',``2'', ``3'', ``4'', or ``5''. My conclusive rating stands at:
\\\hline
As a patron expressing my thoughts, I drafted the ensuing commentary: ``\texttt{[review]}'' \newline
After meticulously assessing my experience, I opted for a star rating from the range of ``1'',``2'', ``3'', ``4'', or ``5''. My definitive rating turned out to be:
\\\hline
As a consumer conveying my perspective,  I authored the following assessment: ``\texttt{[review]}'' \newline
By carefully weighing the aspects of my interaction, I determined a star rating from the possibilities of ``1'',``2'', ``3'', ``4'', or ``5''. My final verdict on the rating is:
\\
\bottomrule
    \end{tabular}
    \caption{Four additional paraphrases of the causal prompt C1 generated with GPT-4.}
    \label{tab:prompts_paraphrases_c1}
\end{table}

\begin{table}[ht]
    \centering \small
    \begin{tabular}{p{7cm}}
\toprule
Prompt Design \\ \midrule
As a customer sharing my experience, I first chose a star rating from the available choices of ``1'',``2'', ``3'', ``4'', or ``5'', and subsequently elaborated on my decision with the following statement: ``\texttt{[review]}'' \newline
The review elucidates the reasoning behind my assigned rating of
\\\hline
As a client providing my opinion, I initially picked a star rating from the range of ``1'' to ``5'', and then proceeded to justify my selection with the following commentary: ``\texttt{[review]}'' \newline
The review sheds light on the rationale for my given rating of
\\\hline
As a patron expressing my thoughts, I started by selecting a star rating from the scale of ``1'' to ``5'', and then offered an explanation for my choice in the following review text: ``\texttt{[review]}'' \newline
The review expounds on the basis for my designated rating of
\\\hline
As a consumer conveying my perspective,  I began by opting for a star rating within the ``1'' to ``5'' spectrum, and then detailed my reasoning in the subsequent review passage: ``\texttt{[review]}'' \newline
The review delineates the grounds for my conferred rating of
\\
\bottomrule
    \end{tabular}
    \caption{Four additional paraphrases of the causal prompt C2 generated with GPT-4.}
    \label{tab:prompts_paraphrases_c2}
\end{table}

\section{Additional Experimental Results}
\subsection{Few-Shot Results}
For reproducibility and controllability, we use the zero-shot prompting setting across the experiments in the main paper, to avoid randomness in few-shot prompting according to which examples are selected as the few shots, and the order of the examples.

As a supplementary information in case this is of some readers' interest, we provide the few-shot prompting results in \cref{tab:prompt_c0_fewshot,tab:prompt_c12_fewshot}.
\begin{table}[ht]
    \centering \small
    \begin{tabular}{llccccccccc}
    \toprule
 && Random 
 & GPT-3 Few-Shot\\ \midrule
\multirow{3}{*}{F1} & Overall & 19.82 {\tiny$\pm$2.07}
& 63.35 {\tiny$\pm$0.80}
    \\
& C1 Subset  & 21.36 {\tiny$\pm$2.26} 
& 54.44 {\tiny$\pm$1.24}  \\
& C2 Subset &  20.43 {\tiny$\pm$2.95} 
& 75.65 {\tiny$\pm$0.45} \\ \hline
\multirow{3}{*}{Accuracy} & Overall &19.78 {\tiny$\pm$2.07} 
& 64.14 {\tiny$\pm$0.86}
\\
& C1 Subset & 20.61 {\tiny$\pm$2.23} 
& 54.22 {\tiny$\pm$1.28} \\
& C2 Subset &  18.86 {\tiny$\pm$2.78} 
& 75.18 {\tiny$\pm$0.53}\\

    \bottomrule
    \end{tabular}
    \caption{Few-shot performance of the standard SA prompts on Yelp-5. We use five paraphrases for the prompt, and report the average performance with the standard deviation.
    }
    \label{tab:prompt_c0_fewshot}
\end{table}
\begin{table}[ht]
    \centering \small
    \setlength\tabcolsep{3.2pt}
    \begin{tabular}{llcccccccccccc}
\toprule
&&Random 
& GPT-3 Few-Shot\\ \midrule
\multirow{4}{*}{F1} 
& Data=C1, Prompt=C1 &20.47 {\tiny$\pm$2.47} 
& 49.18 {\tiny$\pm$0.76} \\
& Data=C1, Prompt=C2 &20.26 {\tiny$\pm$2.31} 
& 52.79 {\tiny$\pm$2.64}\\
& Data=C2, Prompt=C1 & 22.35 {\tiny$\pm$3.02} 
& 80.46 {\tiny$\pm$1.29} \\
& Data=C2, Prompt=C2 & 20.35 {\tiny$\pm$2.18} 
& 75.88 {\tiny$\pm$1.86}\\ \hline

\multirow{4}{*}{Acc}
& Data=C1, Prompt=C1 &19.60 {\tiny$\pm$2.67} 
& 50.12 {\tiny$\pm$0.71}  \\
& Data=C1, Prompt=C2 & 19.68 {\tiny$\pm$2.46} 
& 53.83 {\tiny$\pm$2.46}\\
& Data=C2, Prompt=C1 & 20.97 {\tiny$\pm$3.19} 
& 81.21 {\tiny$\pm$1.17}\\
& Data=C2, Prompt=C2 & 19.03 {\tiny$\pm$2.18}
& 76.36 {\tiny$\pm$1.53}\\
\bottomrule
    \end{tabular}
    \caption{
    Few-shot performance on Yelp using two different causal prompts on the two causal subsets. We use five paraphrases for each prompt, and report the mean performance with the standard deviation.
    }
    \label{tab:prompt_c12_fewshot}
\end{table}

\subsection{$\lambda_1$-$\lambda_2$ Distribution Plot}\label{appd:distr_plot}

\begin{figure}[ht]
    \centering
    \includegraphics[width=0.45\textwidth]{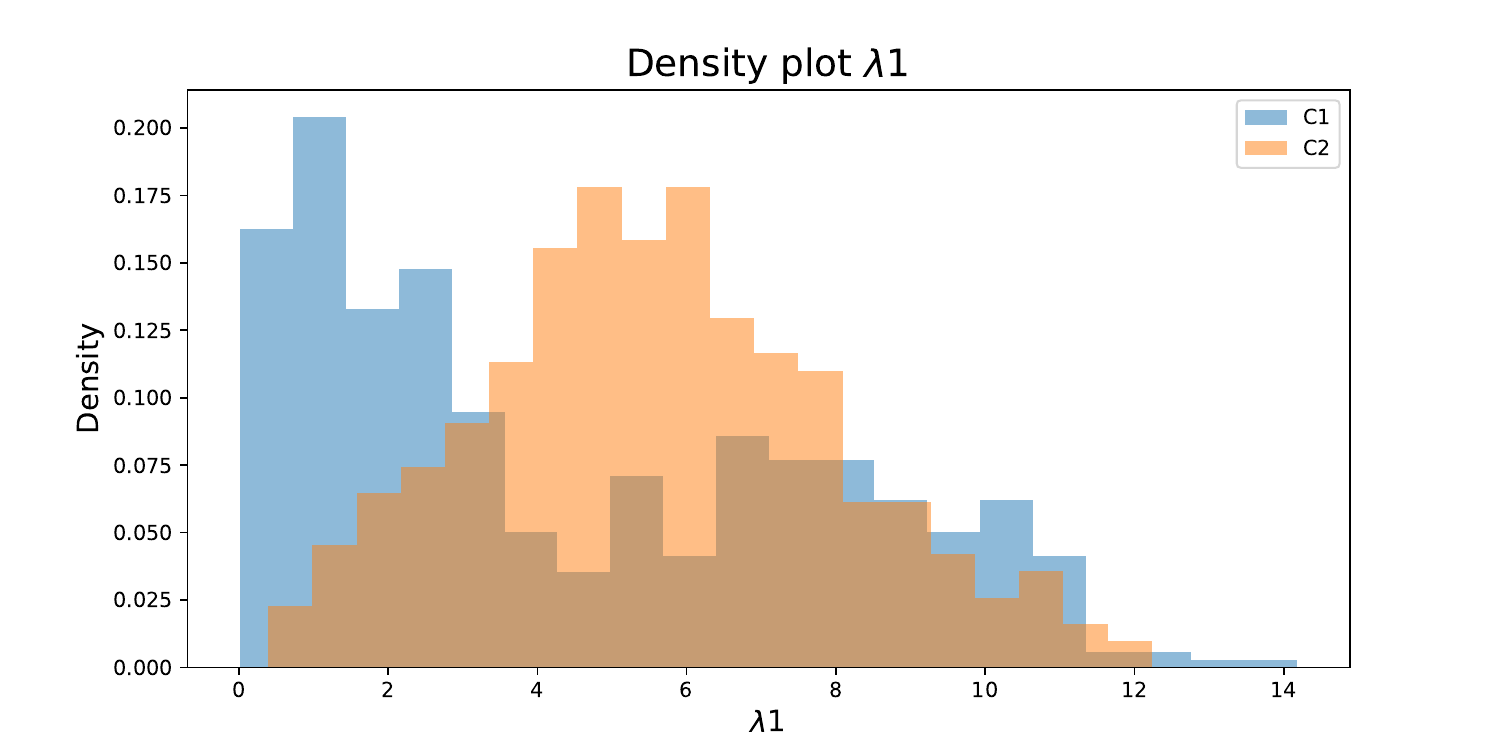}
    \\
    \includegraphics[width=0.45\textwidth]{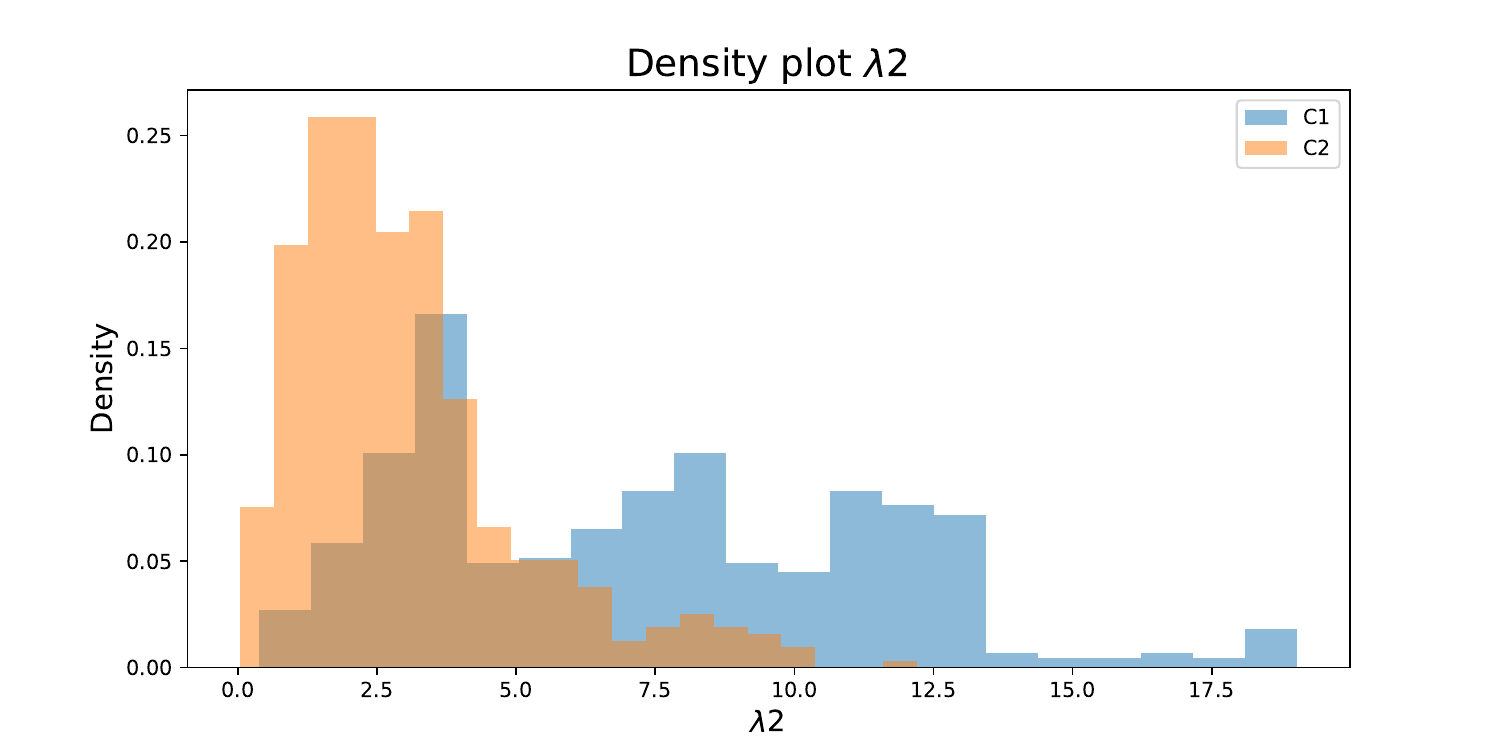}
    \caption{The $\lambda_1$-$\lambda_2$ density plots of C1 (above) and C2 (below). 
    }
    \label{fig:density_lambda}
\end{figure}

To provide a clear understanding of the distributions of \(\lambda_1\) and \(\lambda_2\), we include their density plots of the causal processes C1 and C2 in Figure \ref{fig:density_lambda}.
The mean values of \(\lambda_1\) and \(\lambda_2\) for each group are in \cref{tab:mean}.
\begin{table}[h]
\centering
\begin{tabular}{|c|c|c|}
\hline
 & \textbf{C1} & \textbf{C2} \\ 
\hline
$\mu(\lambda_1)$ & 4.48 & 5.62 \\ 
\hline
$\mu(\lambda_2)$ & 7.31 & 3.02 \\ 
\hline
\end{tabular}
\caption{Mean values of the lambdas for C1 and C2.}
\label{tab:mean}
\end{table}

Further, we performed the Mann-Whitney U rank test to determine if the underlying distributions of \(\lambda_1\) and \(\lambda_2\) for groups C1 and C2 are the same. The results are as follows:
\begin{itemize}
    \item For \(\lambda_1\), the p-value is \(8.4572 \times 10^{-71}\), leading us to reject the null hypothesis that the two groups come from the same distribution.
    \item For \(\lambda_2\), the p-value is \(1.36138 \times 10^{-11}\), also leading us to reject the null hypothesis that the distributions are the same.
\end{itemize}

These statistical results indicate significant differences between the distributions of \(\lambda_1\) and \(\lambda_2\) across the causal process groups, which indicate distinct underlying characteristics in the sentiment dynamics of the two groups.
\begin{figure}[ht]
    \centering
    \includegraphics[width=0.3\textwidth]{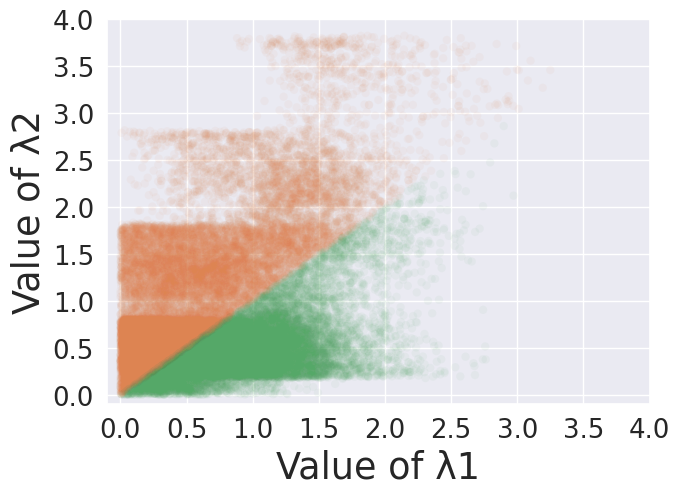}
    \\
    \includegraphics[width=0.3\textwidth]{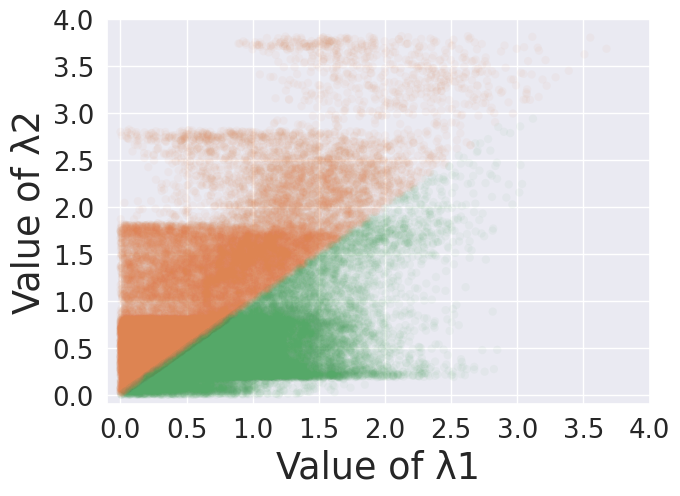}
    \\
    \includegraphics[width=0.3\textwidth]{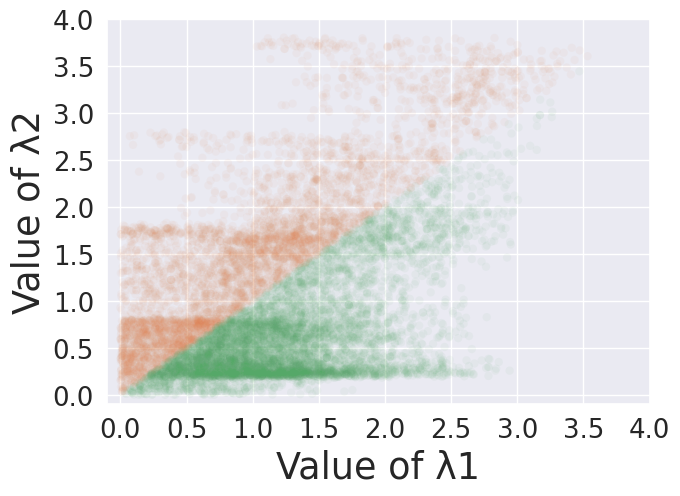}
    \caption{The $\lambda_1$-$\lambda_2$ plot on Yelp-5 (left), Amazon  (middle), and App Review (right). 
    We draw the $y=x$ diagonal line, and 
    the orange dots in the upper-left triangle represent the C1-dominant subset,
    and green dots in the lower-right triangle are the C2-dominant subset.
    }
    \label{fig:lambda_plot}
\end{figure}

\ifarxiv

\section{Emotion Arc Clustering}
We analyze the emotional arc patterns of Yelp reviews. \citet{reagan2016emotional} identified 6 basic emotional arc shapes in stories. However, reviews are usually shorter and therefore present fewer variations. We take each sentence of the review and predict its sentiment. Then we divide the review into ten bins and compute an average sentence sentiment for each decile to make reviews with different lengths comparable. Reviews shorter than 10 sentences generate null values for some deciles which we fill with the information of the next decile.
In \cref{fig:clusters}, we illustrate the 4 clusters found, they have the following characteristics:

\paragraph{Positive + Early Rise:}
This cluster primarily comprises highly positive reviews, where customers express satisfaction and praise for their overall experience. Interestingly, 21.1\% of these reviews begin with a negative first sentence, which often indicates initially low expectations or a negative first impression. However, despite the initial negativity, the reviews tend to turn positive as customers elaborate on their positive experiences.

\paragraph{Negative + Early Fall:}
This cluster mainly consists of predominantly negative reviews. Similarly to the Positive cluster, some reviews (28.14\%) start with a sentence with the opposite sentiment, usually indicating high expectations followed by disappointment.

\paragraph{Rise:}
The main characteristic of this cluster is the positive ending of the review, despite the initial negativity observed in the first half, with an average sentiment of -1.63. An important fraction of the reviews in this cluster (52.49\%) start with a positive comment as a summary, but then proceed to highlight the negative aspects of the experience. Despite the initial criticisms, the reviews conclude with positive points, suggesting that the overall experience was still satisfactory.

\paragraph{Fall:}
In contrast to the previous cluster, the Fall cluster is characterized by a negative ending of the review, despite a generally positive first half with an average sentiment of  2.18. An important proportion (36\%) of the reviews in this cluster begin with a negative comment as a summary, but then proceed to describe the positive aspects before eventually highlighting the negative ones. This cluster showcases a shift in sentiment from positive to negative, indicating a decline in satisfaction as the review progresses.
\begin{table}[t]
    \centering \small
    \begin{tabular}{p{7.2cm}}
    \toprule
    \textit{\textbf{Positive + Early Rise}} \\
    \textbf{Review:} \textit{Was there last Friday. Seats right in front if the stage. The show was good. The headliner, while a bit long, was good. Fantastic service from our waitresses. Will definitely go back.} \\
    \textbf{Review:} \textit{This is by far my favorite Panera location in the Pittsburgh area. Friendly, plenty of room to sit, and good quality food \& coffee. Panera is a great place to hang out and read the news - they even have free WiFi! Try their toasted sandwiches, especially the chicken bacon dijon.} \\
    \hline
    \textit{\textbf{Negative + Early Fall}} \\
    \textbf{Review:} \textit{Pass on this place, there are better restaurants mere feet away.\newline The menu here is too large, which is a sure sign none of the food is going to be good.  And, its not good.  Some of the salads are alright, but its just not good food. \newline The service is friendly and prompt, but the beer is over priced. They do have a good selection though. \newline This place is open late if you need a bite to eat, but there are so much better options out there.} \\
    \textbf{Review:} \textit{Wings are overpriced. And the quality of them are bad. They were tough and greasy. The staff are pleasant but then over all experience was too expensive for a sports bar.} \\
    \hline
    \textit{\textbf{Rise}} \\
    \textbf{Review:} \textit{To be honest, I feel that this is one of the most overpriced restaurants in the entire city. The food is average to good, the place is beautiful with outdoor seating, but in my opinion the price is just not worth it. They have a really good happy hour, so I would definitely recommend going to that and maybe trying an appetizer or two.} \\
    \textbf{Review:} \textit{The first time I came here, I waited in line for 20 minutes.  When it was my turn, I realized I left my wallet in the car.  It hurt so bad, I didn't come back for a year. \newline I can walk to this place from my house- which is dangerous because those biscuits are just OH SO DREAMY.  I can't describe them. Just get some.\newline Do I feel guilty about noshing on fabulous Strawberry Napoleons and Jewish Pizza (kind of like a modified, yet TOTALLY delicious fruitcake bar) at 10:15am?  Hecks, naw... But they do have quiche and some other breakfast-y items for those who prefer a more traditional approach to your stomach's opening ceremony. \newline Just go early :)  They open at 10 on Saturdays.  And bring cash...it's easier that way.} \\
    \hline
    \textit{\textbf{Fall}} \\
    \textbf{Review:} \textit{It's cheap, I'll say that, but otherwise it's bland food served by workers who mostly don't seem to notice they're working, and when they do, only respond snarkily. There are many better vegetarian and vegan options to choose from} \\
    \textbf{Review:} \textit{I do like my Mad Mex, however predictable and non-authentic it may be.  The portion sizes are mammoth and I come away with a satisfied sense of regret.  Their beer menu is happily extensive. Charging me \$9 for chips and salsa is a bit of crime, wouldn't ya say though!?!  I mean, c'mon!  Our service has most times been lacking--a bit rushed and on the inattentive side.  Also, why do you require your wait staff to not servestraws/ lemons/etc unless asked by cusotmers---weirdness-cut out these odd cost-cutting, anti-service friendly measures please} \\
    \bottomrule
    \end{tabular}
    \caption{Example reviews for each emotion arc cluster.}
    \label{tab:cluster_examples}
\end{table}

\begin{figure}[t]
    \centering
    \includegraphics[width=\columnwidth]{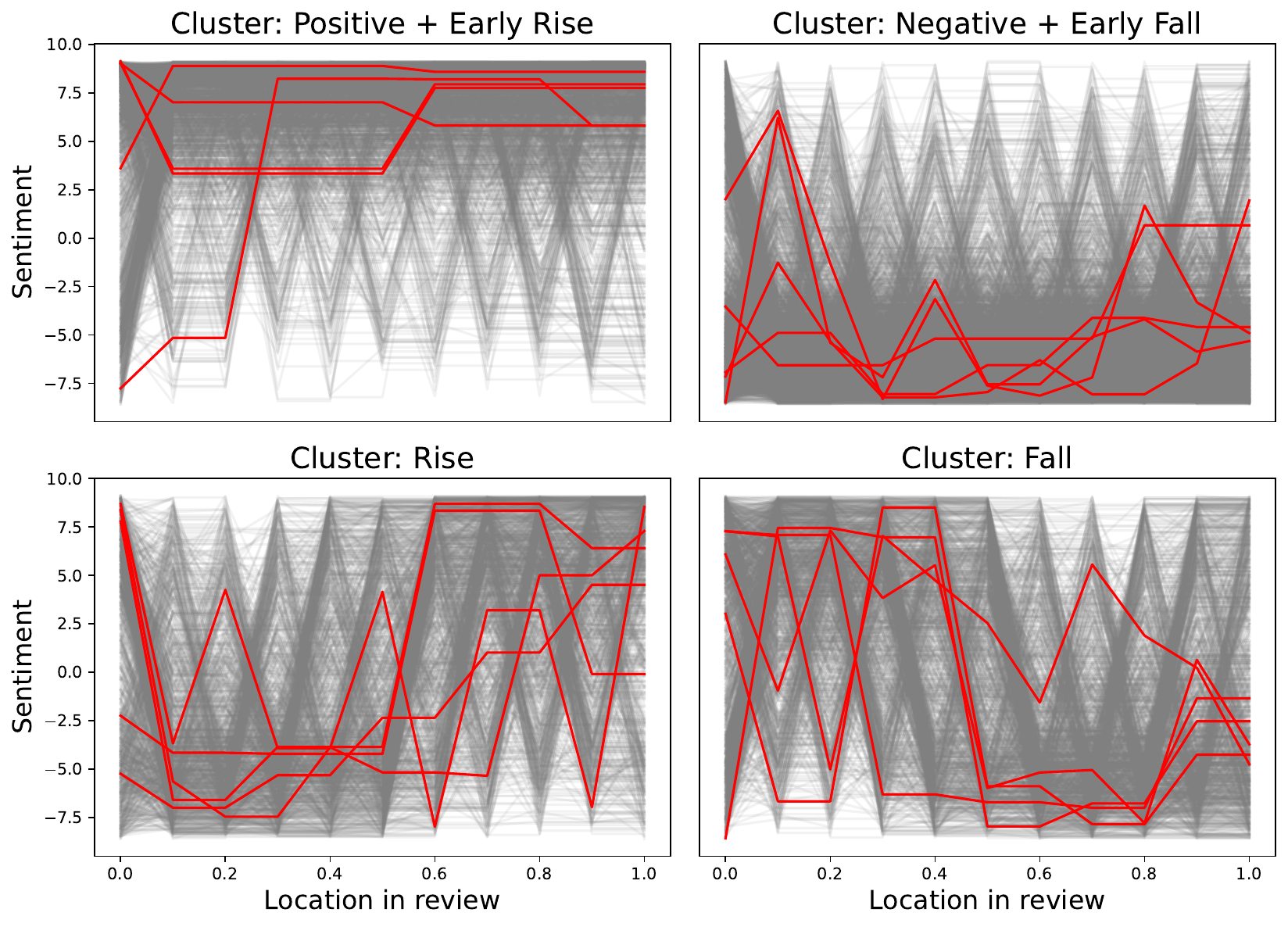}
    \caption{Four emotion arc clusters.}
    \label{fig:clusters}
\end{figure}
\fi
\section{Additional Interpretability by Shapley Values}
\begin{figure*}[t]
    \centering
    \includegraphics[width=\linewidth]{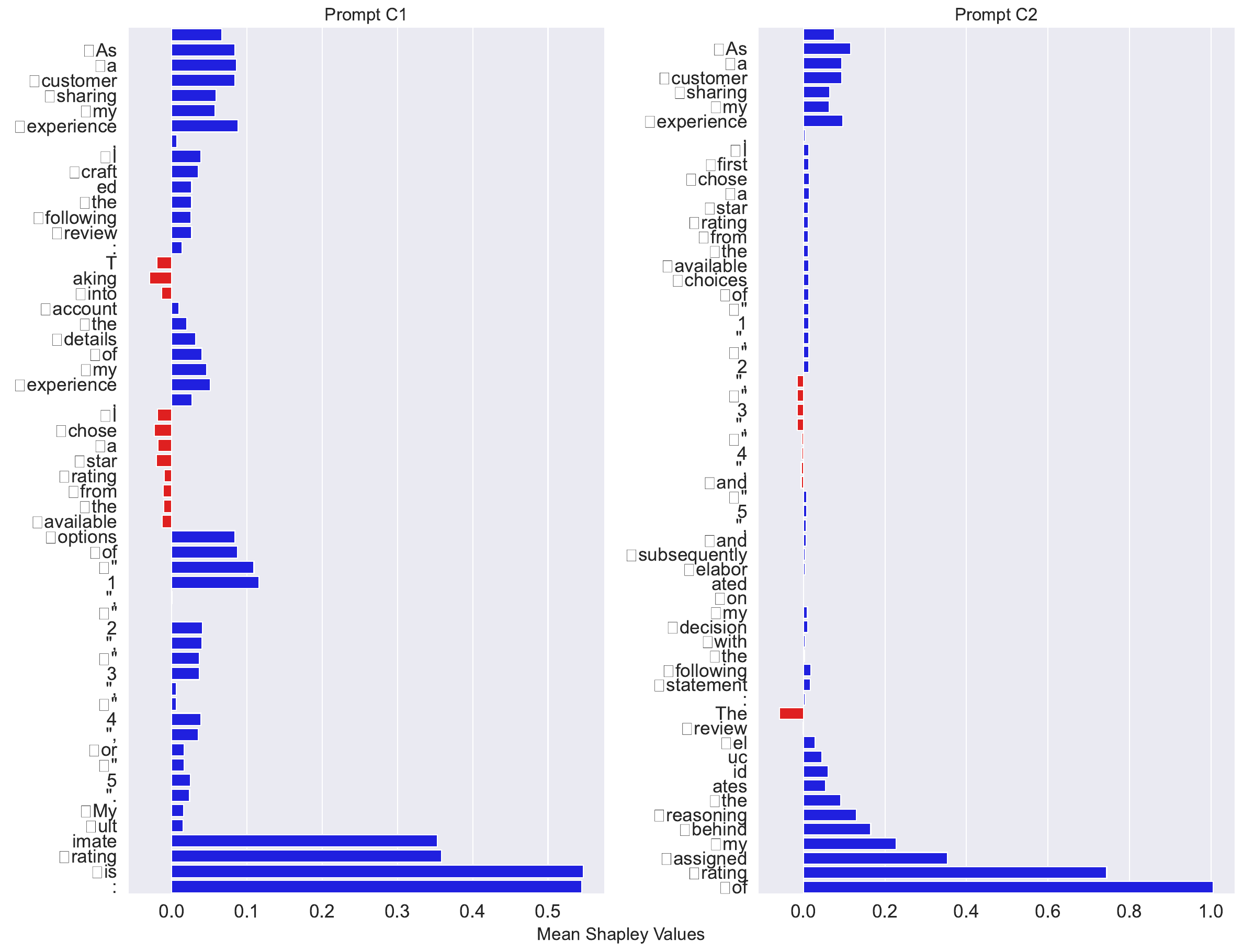}
    \caption{Shapley values for the 2 types of prompt.}
    \label{fig:shapley_values}
\end{figure*}

\begin{figure*}[t]
    \centering
    \includegraphics[width=\linewidth]{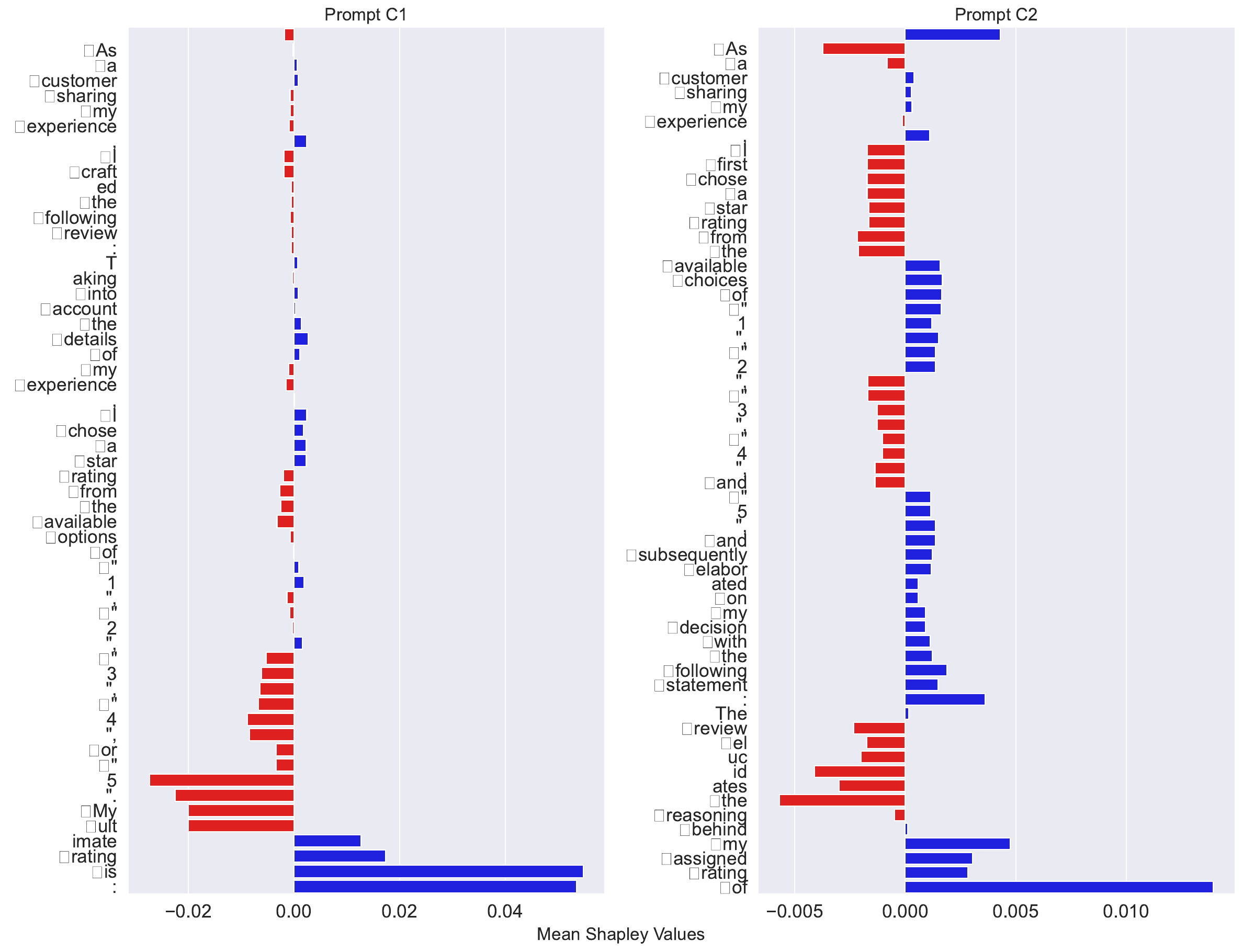}
    \caption{Adjusted shapley values for the 2 types of prompt.}
    \label{fig:shapley_values_norm}
\end{figure*}

We further analyze the effect of each part of the prompts on LLaMa's predictions. Using 50 reviews, we compute the shapley values of each token. In \cref{fig:shapley_values} we observe that the tokens with the largest shapley values are the ones in the end, which is expected since they are the ones helping to form a grammatically correct sentence. To account for that, we subtracted the average shapley values computed for the other possible start rating answers. In \cref{fig:shapley_values_norm} we show the adjusted shapley values. We observe that the tokens in prompt C1 have a larger effect than the tokens in prompt C2. The words introducing the review have a positive effect on C2 but a negative one on C1. Whereas, the phrase ``I chose a star rating'' has a negative effect on C2 but a positive one on C1.

\end{document}